# FASE-AL — Adaptation of Fast Adaptive Stacking of Ensembles for Supporting Active Learning

Agustín Alejandro Ortiz-Díaz[1], Fabiano Baldo[1], Laura María Palomino Mariño[2]
and Alberto Verdecia Cabrera[3]

[1]Santa Catarina State University, Joinville, Santa Catarina, Brazil
agaldior@gmail.com and fabiano.baldo@udesc.br
[2]Pernambuco Federal University, Recife, Pernambuco, Brazil
lmpm@cin.ufpe.br
[3]Granma University, Manzanillo, Granma, Cuba
averdeciac@udg.co.cu

## Abstract

*Classification algorithms to mine data stream have been extensively studied in recent years. However, a lot of these algorithms are designed for supervised learning which requires labeled instances. Nevertheless, the labeling of the data is costly and time-consuming. Because of this, alternative learning paradigms have been proposed to reduce the cost of the labeling process without significant loss of model performance. Active learning is one of these paradigms, whose main objective is to build classification models that request the lowest possible number of labeled examples achieving adequate levels of accuracy. Therefore, this work presents the FASE-AL algorithm which induces classification models with non-labeled instances using Active Learning. FASE-AL is based on the algorithm Fast Adaptive Stacking of Ensembles (FASE). FASE is an ensemble algorithm that detects and adapts the model when the input data stream has concept drift. FASE-AL was compared with four different strategies of active learning found in the literature. Real and synthetic databases were used in the experiments. The algorithm achieves promising results in terms of the percentage of correctly classified instances.*

## 1. Introduction

In the last years, special attention has been given to classification algorithms. In particular, the algorithms that manipulate data stream in the presence of concept drift have been deepened. Several of the early scientific investigations focused mainly on supervised learning. However, in many practical situations of supervised learning, labeling instances is often an extremely expensive task. Some of these scenarios are the classification of web pages, the detection of spam emails and the detection of network fraud. Classifying instances in these scenarios may require a lot of time for the experts in each area [1].

For the analysis of data stream incrementally, the classification task is usually performed in a sequence of instances $S = e_1, e_2, \ldots, e_j, \ldots$ arriving over time. Each training instance $e_j = (\vec{x}_j, y_j)$ is formed by a vector $\vec{x}_j$ and a discrete classification value $y_j$, which is taken from a finite set Y named classes. Each vector $\vec{x}_j \in \vec{X}$ has the same dimensions. It is assumed that there is an underlying function $y = f(\vec{x}_j)$ and the main objective is to construct a model from S that approximates $f$ as $\hat{f}$ in order to predict the class of unlabeled instances, so that $\hat{f}$ maximizes the prediction accuracy [2].

On the other hand, concept drift is categorized into two types according to the number of instances that delay the transition from one concept to another. A change is considered gradual when the transition period between two concepts contains a certain number of instances and abrupt when the transition between consecutive concepts is instantaneous [3, 4].

Inducing general models from data is much more difficult when the instances come without labels. However, due to the time consuming and the resources expended, most of the instances of a stream generated every day are not labeled. To face this practical problem, in recent years researchers focused on propose learning paradigms that could overcome this shortcoming. They pursue paradigms whose main objective is to maintain a high performance of the models generated from a reduced number of instances with labels. Two of the paradigms with the best results are semi-supervised learning [5, 6] and active learning [7, 19]. The first of these paradigms works directly on instances without labels. It tries to group the examples by their similar characteristics from the analysis of their possible underlying distributions. The second one selects the instances that provide the most significant information to be labeled and used to train the model.

Adaptive Stacking of Ensembles (FASE) [8] is an ensemble algorithm designed only for supervised learning. That is, it builds models by learning from a data stream where all the instances are labeled. This algorithm is able to detect and adapt to the concept drift in the input data stream, whether abrupt or gradual. This ensemble has a drift detection method inserted. One of the advantages of incorporating this mechanism is to exploit the capacity of the ensembles to adapt to the gradual changes, combined with the natural mode of operation of the drift detection method during the abrupt changes [2]. FASE has a fixed amount of base learners introduced by parameters. This last characteristic of the model allows the use of a meta-learner who learns from the intermediate results provided by the base learners. This meta-learner combines the partial results returning a general prediction. This paper proposes an adaptation of the FASE algorithm that maintains its main characteristics but enabling it to process data streams where only a small percentage of the data is labeled. In order to reduce the cost and time to obtain the class of unlabeled data, we adapted the FASE algorithm to the paradigm of active learning.

The paper is organized as follows. Section 2 covers the works related to the classification of stream data with ensembles and active learning. Section 3 presents the active learning algorithm adaptation for drifting data streams. In section 4 a spatial and temporal study of the complexity of the proposal is made. Section 5 describes the assessment of the results of comparisons made with other strategies for active learning. Section 6 summarizes the conclusion and future works.

## 2. RELATED WORK

### 2.1. Concept drifts in data streams

In general, to address the problem of concept drift two types of strategies are defined [9]; strategies in which the learning adapts in a periodic time interval, regardless there is a concept drift or not; and strategies in which the concept drift is first detected, then the learning adapts according to the change.

The ensembles are usually included within the first strategy. This type of model has intrinsic mechanisms that allow it to evolve in a regular way without having to directly detect the concept drift. However, recent researches propose incorporating change detectors within the ensembles. The incorporation of a change detector in the model can enable the ensemble to gradually adapt itself to the changes taking the advantage of using the natural mode of operation of the detectors during the abrupt changes [2].

Among the first proposals of ensemble algorithms, the following can be mentioned: Streaming Ensemble Algorithm (SEA) [10], Fast Adapting Ensemble (FAE) [2], and Fast Adaptive Stacking of Ensembles (FASE) [8]. In addition, other methods were developed for the online detection of changes in distribution, among these are: Drift Detection Method (DDM) [9], Early Drift Detection Method (EDDM) [11], and Hoeffding-based Drift Detection Method (HDDMA-test) [12].

## 2.2. Active learning

Many real-life situations generate instances without labeling. The process of labeling them can lead to excessive expend of resources. Due to this problem, in recent years several learning paradigms have been proposed aiming at reducing the cost of labeling without significantly compromising the performance of the model. Among the main paradigms that follow this objective, it can be found the semi-supervised learning [6] and active learning [7]. Active learning focuses primarily on maintaining high levels of correct predictions from a limited number of labeled instances. In general, the strategy followed to achieve this purpose is to use a function that selects the instances with a greater load of significant information. The next step is to obtain the real class of these instances and use them in the training process of the model [1].

To construct this selection function, the most commonly used idea is to apply the probabilistic vector of the predictions of each of the possible classes. Many works propose to use an uncertainty function to assess the significance of an instance, where the most valuable examples are selected to be labeled and used to train the algorithm [14].

The Entropy ($EM$) metric is a measure of information retrieval that represents the uncertainty taking into account all the probabilistic vector components of the predictions for each class. For this reason, it is usually one of the most used functions. Given a prediction hypothesis $\Theta$ of an instance $\vec{x}_j$ the uncertainty can be calculated as (1) [1]:

$$EM = -\sum_{j} P\Theta(\hat{y}_i | \vec{x}_j) \log P\Theta(\hat{y}_i | \vec{x}_j) \tag{1}$$

Where $\hat{y}_i$ represents posterior probability of the instance $\vec{x}_j$ being a member of the *ith* class, which takes into account all possible labels. Other variants of this same main entropy formula have been used in other investigations [15].

In [16], the authors introduce a framework specialized in the active learning paradigm, the manipulation of data streams and the treatment of concept drift. Three strategies are proposed in this paper.

i. Random Strategy uses a uniform random variable $£ \in [0,1]$ to select the instances that will be labeled. The instance is selected if $£ \leq B$ is a probabilistic budget initially established.

ii. On the other hand, the objective of the Variable Uncertainty strategy is to select the instances uniformly over time. To achieve this a threshold is used, which is adjusted as time progresses. The threshold expands if the uncertainty is low; and on the contrary, the threshold is contracted if the uncertainty is high.

iii. By last, Split Strategy combines the Random Strategy and the Variable Uncertainty Strategy. This strategy divides the input data into two streams.

Without losing the generality, the Random strategy is applied to the first stream and the Variable Uncertainty Strategy to the second stream. Both sequences are used to train the classifier, but only the first one is used to detect the concept drift.

Also, in [16], three indispensable requirements are defined to construct strategies for a labeling process that efficiently manipulates the concept drift. (i) The available budget should never be exceeded. That is, the balance of the examples labeled overtime must be maintained. (ii) To detect any type of concept drift, it must be guaranteed that all the space of the instances must be analyzed. (iii) It must be ensured that the initial distribution of input data is preserved.

## 2.3. FASE: Fast adaptive stacking of ensembles

The FASE algorithm is the starting point for the proposal of this paper. FASE is an ensemble designed to learn from non-stationary data streams. Only two parameters are necessary to configure it [8]; the confidence level, used in the strategy to detect changes; and the base learner number, fixed quantity of basic classifiers that will be part of the ensemble.

FASE associates methods of detection of changes to each of its base learners. That is, it actively manages the concept drift. First, the concept drift is detected and then takes actions to modify the old learning models. FASE uses the model proposed in [12] (HDDMA-test) as a change detector. HDDMA-test offers mathematical guarantees for false positive and false negative rates. FASE has a fixed number of base classifiers, and it uses a meta-learner to unify the intermediate predictions [8]. Its architecture can be seen in Figure 1.

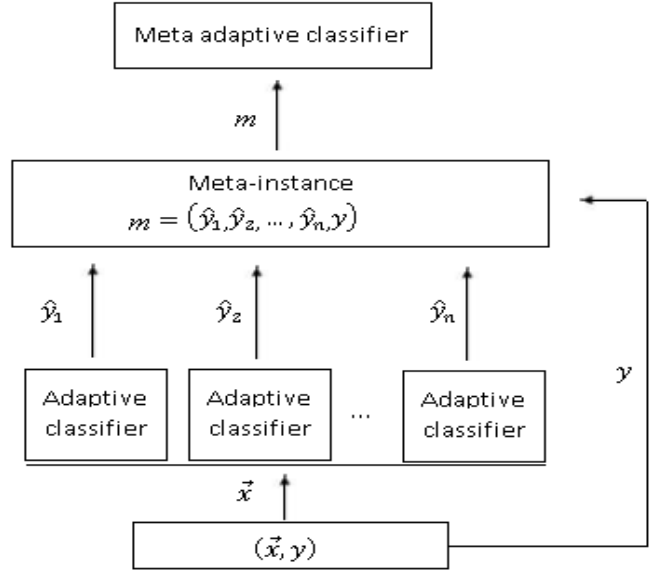

Figure 1. Scheme of FASE algorithm

FASE constructs meta-instances to train its meta-learner using a test-then-train proposal [13]. Thus, for each example of training $e_j = (\vec{x}_j, y_j)$ from the original stream a meta-instance $m_j = (\hat{y}_1, \hat{y}_2,...,\hat{y}_n, y)$ is generated. Each attribute value $\hat{y}_n$ of the meta-instance $m_j$ corresponds to the prediction from the base classifier $i$ for the original training instance. For this meta-instance $m_j$ he value $\bar{y}_i$ is the class label predicted by the base classifier. Each meta-instance $m_j$ has as class the same class that accompanied the original training instance [8].

## 3. FASE-AL: FAST ADAPTIVE STACKING OF ENSEMBLES WITH ACTIVE LEARNING

FASE-AL is an adaptation of the FASE algorithm to work under the paradigm of active learning. The new algorithm has the ability to perform in a scenario where only a part of the training instances is labeled. To achieve this, a variant of the **"Split Strategy"** has been added to the meta-instance level, inside the **"Selection Strategy"** box, as showed in Figure 2. The "Split Strategy" was selected because it has important characteristics described in section 2.2.

Analyzing figures 2 and 3 we can see that the FASE and FASE-AL algorithms basically differ in the strategy of selection of instances. This strategy is responsible for processing and selecting unlabeled instances. In this way, more relevant information is incorporated into the induced model. A characteristic of the FASE-Al algorithm is that it incorporates the selection strategy at the level of the meta-instances. This feature differentiates it from other models of the active learning paradigm.

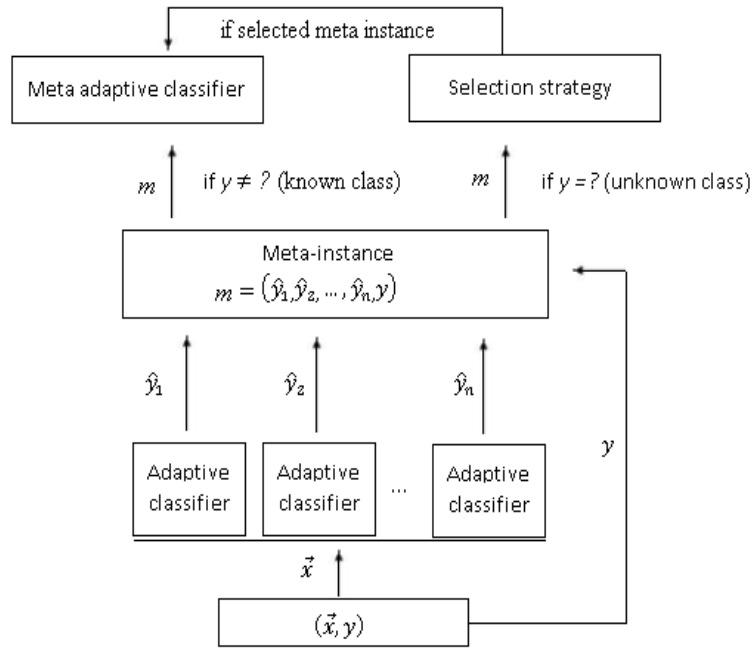

Figure 2. Scheme of FASE-AL algorithm

Concerning the original Split Strategy [16] mentioned in section 2.2, a small modification has been added by replacing the Least Confident Metric to the Entropy Metric. This was done due to the fact that, according to the authors of [1], the Entropy Metric is a measure of information retrieval that represents the uncertainty taking into account all the components of the probabilistic vector of the class predictions. Unlike the Least Confident Metric which only takes into account the maximum posterior probability component. Therefore, in lines 2 and 3 of Algorithm 2, the Least Confident Metric was replaced by the Entropy Metric formula (*EM*: Formula 1).

The vector $\bar{y}_t$ (line 2 Algorithm 2) is formed with the values of probabilities that each class is predicted. This vector is calculated from the meta-instance and it is used for the calculation of entropy (lines 2 and 3 Algorithm 2). As explained in the FASE description, the meta-instance is formed by each of the classes predicted by the base classifier. Then it is possible to calculate the probability percentage for each of the classes. These percentages constitute the vector whose dimension is equal to the total number of classes. Algorithms 1 and 3 are exactly those described in [16]. However, Algorithm 2 has the modification described above.

---

**Algorithm 1:** SPLIT
**Require:** incoming example $X_t$, trained classifier L,
        threshold adjustment step s ∈ (0,1],
        proportion of random labeling δ ∈ (0,1), budget B
**Ensure:** labeling ∈ {true, false}
**1: Initialize:** labeling threshold θ ← 0
  and store the latest value during operation
**2: if** η < δ, where η ~ U [0,1] is random **then**
**3:**    **if** change detected then
**4:**        cut the training window
**5:**    **end if**
**6:**    **return** (labeling ← RANDOM (B))
**7: else**
**8:**    **return** (labeling ← VAR_UNCERTAINTY ($X_t$, L, s))
**9: end if**

---

**Algorithm 2:** VAR_UNCERTAINTY
**Require:** incoming example *Xt*, trained classifier *L*,
        threshold adjustment step s ∈ (0,1]
**Ensure:** labeling ∈ {true, false}
**1: Initialize:** labeling threshold θ ← 1
**2:** Calculate $\vec{y}_t$ **and** *EM* // vector of the probabilities of each class
  and entropy. It is obtained from *Xt* (meta-instance)
**3: if** $1 - EM < \theta$ **then**
    //uncertainty below the threshold
**4:**    decrease the uncertainty region θ ← θ (1 − s);
**5:**    **return** (labeling ← true)
**6: else**
    //certainty is good
**7:**    uncertainty region wider θ ← θ (1 + s);
**8:**    **return** (labeling ← false)
**9: end if**

---

**Algorithm 3:** RANDOM
**Require:** $X_t$: incoming example, B: budget
**Ensure:** labeling ∈ {true, false}
**1:** generate a uniform random variable ξ ∈ [0,1];
**2: return** (labeling ← true (ξ ≤ B))

---

## 4. SPATIAL AND TEMPORAL COMPLEXITY ANALYSIS

To perform a spatial and temporal study is particularly more relevant when the learning process is done online in a data stream, where it is possible to have non-ending datasets. In this paper, the complexity analysis of the proposed algorithm was done using the Naive Bayes (NB) as base learner. To identify the complexity using other base learners a similar study should be done.

In terms of spatial complexity, it will depend mainly on the number $n$ of base classifiers that belonged to the ensemble. The theoretical space complexity for NB algorithm is $O(n_{attr} \cdot n_{value} \cdot n_{class})$, where $n_{attr}$ is the number of attribute, $n_{value}$ is the number of values per attribute and $n_{class}$ is the alternative values for the class [17]. If we assume that the ensemble is defined by a finite number of base learners $n$ the spatial complexity of FASE-AL algorithm is $O(n \cdot n_{class} \cdot n_{attr} \cdot n_{value})$.

Regarding the temporal complexity, as FASE-AL is tailored for continuous running on a data stream, we will measure the running time for each example. Therefore, to carry out the analysis, three different situations must be taken into account.

i. Create a new base learner. The temporal complexity depends on the temporal complexity of the selected base classifier. In this case, NB requires constant time to process each example. The theoretical time complexity for NB classifier is $O(N \cdot n_{attr})$ where $N$ is the number of training examples and $n_{attr}$ is the number of attributes of each example [17], that for an example is, $O(n_{attr})$ (N=1).

ii. Update the base learners. In the updating process, each example must be tested with each base learner in order to detect the concept drift. The theoretical time complexity for Hoeffding Drift Detection Method (HDDMA-test) [12] is O(1). Then, in the worst-case scenario the temporal complexity is O(n· $n$attr), where n is the number of base learners.

iii. Update the meta-learner. The meta adaptive learner is also an NB algorithm. It is trained with the meta-instances. These meta-instances have as many attributes as learners in the ensemble. Therefore, the temporal complexity of meta adaptive learner is $O(n)$. The time complexity of the strategy of selection of instances whose label is not known only depends on the number of attributes of the meta-instance. Therefore, as this number coincides with the number of base learners from the ensemble, its temporal complexity is $O(n)$.

At this point, it is important to highlight that the time complexity of the algorithms within the active learning paradigm will depend to a great extent on the time required to obtain the real class of the selected instances. This, in turn, will greatly depend on the type of problem. Assuming that the complexity of obtaining the real class of each instance unlabeled is $O(1)$, then the temporal complexity of the FASE-AL algorithm is $O(n \cdot n_{attr})$.

## 5. EXPERIMENTAL STUDY AND RESULTS ASSESSMENT

The main objective of the following experimental study is to verify that the FASE-AL algorithm obtains promising results in a semi-supervised learning environment. In addition, check that FASE-AL maintains the ability of FASE to detect and adapt to possible concepts drift, both abrupt and gradual.

All the experiments described below were performed in the Massive Online Analysis framework (MOA) [18]. MOA is a framework specialized in the mining of data streams. This framework groups a set of learning algorithms, evaluation tools and generators of data streams. In this section, the FASE-AL was compared with four different strategies of active learning found implemented in MOA. These four strategies are adaptations of the three strategies that were proposed in [16], already described in previous sections. The four strategies for active learning were implemented within a general model called *AL-Uncertainty*.

## 5.1. Datasets and algorithms configuration

Table 1 shows the main characteristics of the datasets used in the experiments. These ones were selected trying to guarantee a high diversity in terms of types of attributes, number of attributes, number of classes and number of instances. Both artificial and real datasets were used to perform the experiments. Within the artificial data sets were included characteristics such as abrupt changes, gradual changes and artificial noise.

The experiments were divided into four different scenarios in order to check the performance of the algorithms in different situations. For the first scenario, five synthetic databases were generated. In this scenario, the concept is always stable (there is no concept drift). For scenarios two and three, five synthetic databases were also generated. In each database of the second scenario, abrupt concept drifts were inserted and in the third scenario, gradual concept drifts were inserted. All the synthetic databases were built with 1000000 instances, Table 1.A. In addition, each one has a 10 % noise inserted randomly; this is a MOA framework option to achieve a greater similarity with real situations. In the fourth scenario, the five real databases shown in Table 1.B were used.

Table 1: Principal characteristics of the datasets used in the experiments.

| Dataset | | Attributes | | Number of | |
|---|---|---|---|---|---|
| **Name** | **Acronym** | **Numeric** | **Nominal** | **Instances** | **Classes** |
| | | **A. Artificial datasets** | | | |
| **SEA** | SEA | 3 | 0 | 1000000 | 2 |
| **STAGGER** | STA | 0 | 3 | 1000000 | 2 |
| **LED Display** | LED | 0 | 24 | 1000000 | 10 |
| **Agrawal** | AGR | 6 | 3 | 1000000 | 2 |
| **Hyperplane** | HYP | 10 | 0 | 1000000 | 2 |
| | | **B. Real datasets** | | | |
| **Spam** | SPA | 0 | 500 | 9323 | 2 |
| **Weather** | WEA | 8 | 0 | 18159 | 2 |
| **Electricity** | ELE | 7 | 1 | 45312 | 2 |
| **Connect-4** | CON | 0 | 21 | 67557 | 3 |
| **Poker Hand** | POK | 0 | 10 | 1000000 | 10 |

FASE-AL was configured to use the Naive Bayes (*NB*) algorithm as both base classifier and the meta-learner. The significant level $1-\lambda$ of *HDDMA-test* was configured to $\lambda = 0.005$ for the warning status and to $\lambda = 0.001$ for the drift status. The number of base classifiers was configured to 10. The values of the parameters for the selection strategy were: $s = 0.01$ (recommended value); $\delta = 0.05$; and $B = 0.05$.

The four strategies for active learning were implemented within a general model called *AL-Uncertainty*. The four strategies are recognized with the following names in MOA: *Fixed-Uncertainty*, *Var-Uncertainty*, *Rand-Uncertainty*, and *Sel-Sampling*.

The values of the main parameters were adjusted to: *budget = 0.1*; fixed *threshold = 0.9* and *step = 0.01*. The rest of the parameters took their values for defects. All the algorithms were evaluated by an *Active-Learning-Prequential-Evaluation-Task*. This evaluation strategy was implemented in MOA within the environment for active learning to evaluate the algorithms following the test-then-train strategy.

## 5.2. Results and discussions

Tables 2 and 3 show the results of the experiments obtained from the four scenarios. Each of the algorithms was applied 10 times to each synthetic and real dataset. Two averages appear in each cell. The first value is the percentage of correctly classified instances (accuracy) and the second value is the execution time in seconds (runtime).

Table 2: Percentage of correctly classified instances (accuracy) and runtime of the studied algorithms on the synthetic datasets. A) Synthetic datasets without concept drift. B) Synthetic datasets with abrupt concept drifts. C) Synthetic datasets with gradual concept drift.

| Bases/ Algorithms | SEA | STA | LED | AGR | HYP |
|---|---|---|---|---|---|
| A) Without Concept Drift ||||||
| **FASE-AL** | **87,84** 13,65 | **100** 5,85 | **74,00** 20,59 | 93,73 8,79 | 94,14 6,69 |
| **Fixed-Uncertainty** | 64,32 12,79 | 99,99 4,53 | 9,97 **18,08** | 49,66 4,57 | 93,55 **3,28** |
| **Var-Uncertainty** | 86,81 **12,40** | 99,98 4,83 | 73,89 19,66 | **94,05** **2,96** | **94,22** 3,53 |
| **Rand-Uncertainty** | 87,68 12,95 | 99,99 **4,72** | 73,88 19,83 | 92,25 3,17 | 94,16 3,45 |
| **Sel-Sampling** | 87,47 12,76 | **100** 4,83 | 73,83 19,56 | 92,83 3,78 | 94,39 3,57 |
| B) Abrupt Concept Drift ||||||
| **FASE-AL** | **87,27** 14,06 | **100** 6,21 | **74,22** 22,15 | 81,97 27,32 | **94,63** 36,13 |
| **Fixed-Uncertainty** | 61,04 **11,68** | 65,76 3,25 | 10,06 17,71 | 48,48 24,12 | 49,82 **32,74** |
| **Var-Uncertainty** | 86,37 12,97 | **100** **3,06** | 73,72 **17,15** | 81,47 23,40 | 94,37 34,01 |
| **Rand-Uncertainty** | 85,14 12,03 | **100** 3,11 | 73,71 18,26 | 79,79 **23,39** | 94,02 33,99 |
| **Sel-Sampling** | 86,59 12,11 | **100** 3,25 | 73,70 17,76 | **83,84** 23,48 | 94,52 34,12 |
| C) Gradual Concept Drift ||||||
| **FASE-AL** | **88,95** 14,15 | **99,70** 6,32 | **73,91** 21,33 | 81,94 27,74 | 94,52 36,13 |
| **Fixed-Uncertainty** | 61,13 12,93 | 65,773 ,28 | 10,05 17,82 | 48,47 24,07 | 49,80 **32,73** |
| **Var-Uncertainty** | 86,92 12,87 | 99,61 **3,12** | 73,42 **17,11** | 79,12 23,34 | 94,41 33,06 |
| **Rand-Uncertainty** | 83,20 **12,03** | 99,68 3,17 | 72,96 **17,11** | 78,17 **23,31** | 94,05 34,04 |
| **Sel-Sampling** | 87,86 12,93 | 99,683 ,25 | 72,91 19,51 | **83,67** 23,49 | **94,53** 34,12 |

Table 2.A shows the results achieved by the algorithms in the first scenario (synthetic datasets without change). Empirically we can verify that the algorithms *Var-Uncertainty* and *FASE-AL* have the most promising results in terms of the percentage of correctly classified instances. The results of the rest of the algorithms can be considered inferior. These three other algorithms never get the best accuracy.

Table 2.B and Table 2.C show the results achieved by the algorithms where abrupt and gradual concept drifts have been inserted, respectively. Each of these synthetic datasets has three points of change inserted. The first point is always around the 250000 instances, the second is around the 500000 instances and the third is around the 750000 instances. The datasets of the second experimentation scenario have inserted abrupt concepts drifts (0 transition instances from one concept to another). The datasets of the third scenario have inserted gradual concept drifts (1000 transition instances from one concept to another).

In the last two scenarios analyzed (Table 2.B and Table 2.C), we can establish empirically that the FASE-AL algorithm has the most promising results in terms of the percentage of correctly classified instances. In 7 of the 10 cases studied, this algorithm obtains the highest values in comparison with the results of the other algorithms.

The following four figures show the particular accuracy results of the algorithms on the LED data set with gradual changes. As described earlier, this dataset has three points of concept drift inserted. The first change occurs around instance 250000, the second change around instance 500000 and the last change around instance 750000. When analyzing Figures 3-6 we can be observed how all algorithms have a decrease in the percentage values of accuracy in the environment of the points of the concept drift. This result is expected because a new concept begins to arrive. However, we can see that this fact occurs less markedly for the FASE-AL algorithm.

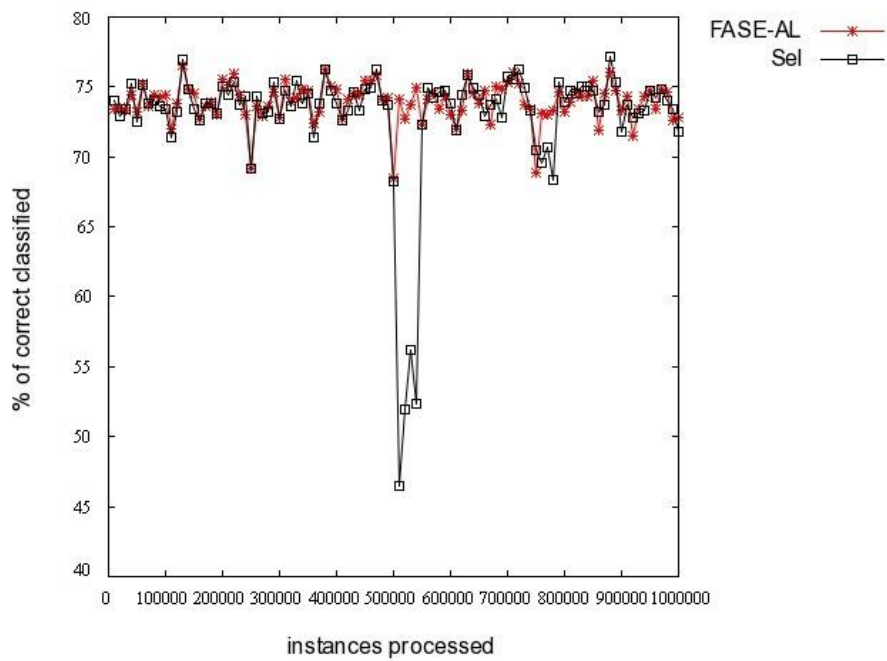

Figure 3. Scheme of FASE-AL algorithm Percentage of correctly classified instances of the algorithms FASE-AL and Sel-Sampling on LED dataset with gradual changes. Concept drifts occur every 250.000 instances.

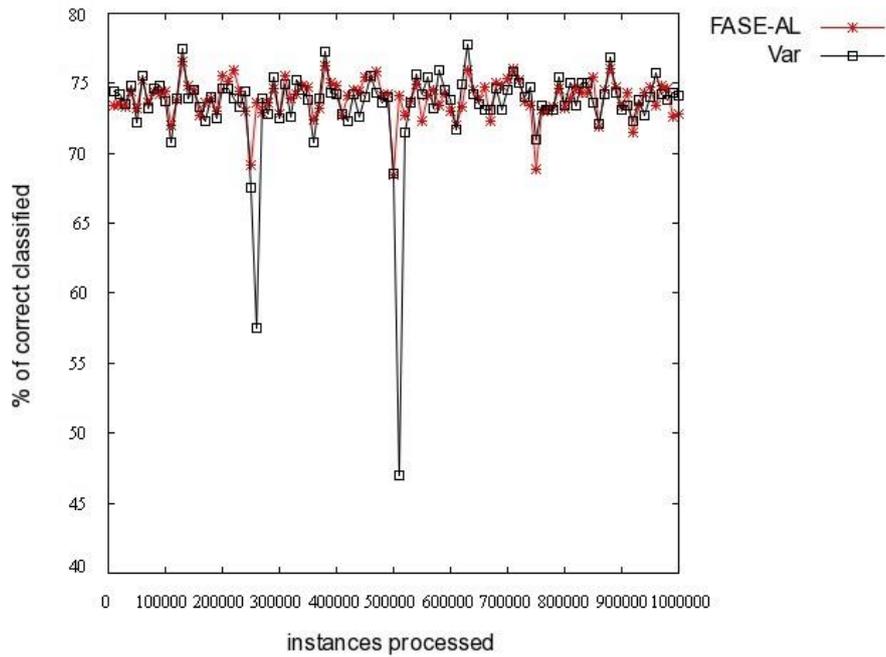

Figure 4. Scheme of FASE-AL algorithm Percentage of correctly classified instances of the algorithms FASE-AL and Var-Uncertainty on LED dataset with gradual changes. Concept drifts occur every 250.000 instances.

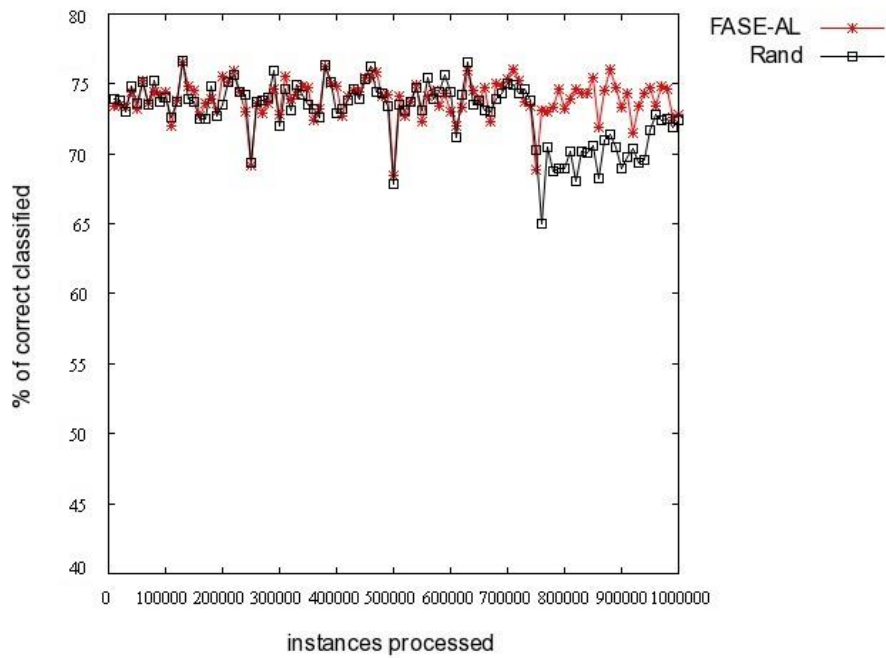

Figure 5. Scheme of FASE-AL algorithm Percentage of correctly classified instances of the algorithms FASE-AL and Rand-Uncertainty on LED dataset with gradual changes. Concept drifts occur every 250.000 instances.

In Figure 3 we can see how the Sel-Sampling algorithm markedly lowers its accuracy values when analyzing the instances in the vicinity of the change points 2 and 3; that is, around instances 500000 and 750000. A similar result is shown in Figure 4 for the Var-Uncertainty algorithm. This algorithm lowers its accuracy values around changes points 1 and 2; that is, around instances 250000 and 500000. On the other hand, Figure 5 shows that the Rand-Uncertainty algorithm fails to recover its accuracy values after the last point of change, that is, after instance 750000. Finally, it is necessary to highlight that the fixed-uncertainty algorithm obtained very bad results on the LED data set, Figure 6.

Similar results of the algorithms could be observed, regarding the accuracy values, with respect to the rest of the synthetic data sets. This happens both in scenarios with abrupt changes, as well as in scenarios with gradual changes.

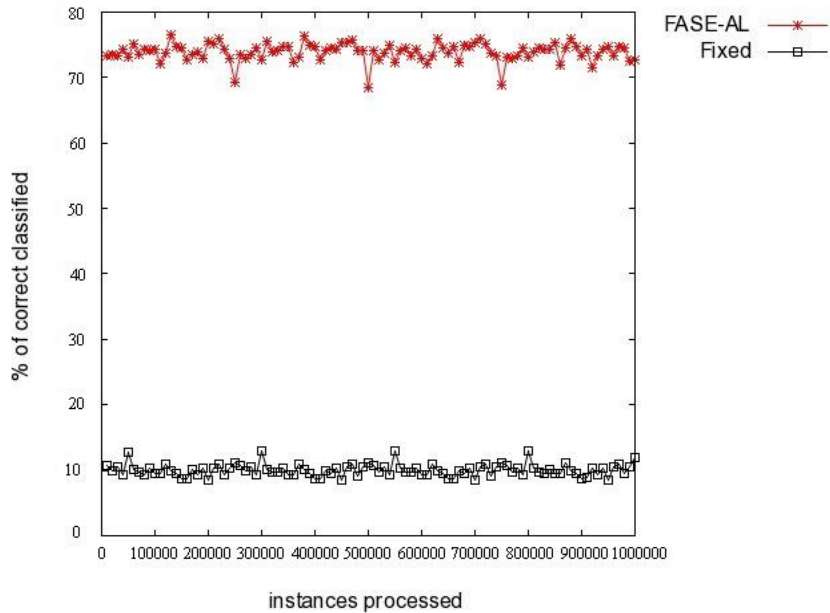

Figure 6. Scheme of FASE-AL algorithm Percentage of correctly classified instances of the algorithms FASE-AL and Fixed-Uncertainty on LED dataset with gradual changes. Concept drifts occur every 250.000 instances.

Table 3 shows the performance of the five algorithms assessed with real datasets. Once again, we can establish that the results of the FASE-Al algorithm are the most promising. In four of the five cases studied, FASE-AL had the best results.

The results of the *FASE-AL* algorithm were compared with the results of the rest of the algorithms. This comparison was made using the Wilcoxon signed-rank test. The test showed a significant difference with the algorithms *FixedUncertainty*, *VarUncertainty* and *RandUncertainty*. However, it did not provide sufficient evidence to affirm a significant difference with the *SelSamplingL* algorithm.

The main deficiencies of the system to classify data streams were observed in two fundamental points. First, the results related to the runtime. The FASE-AL algorithm has lower values in the experiments. The other four algorithms obtain similar results in terms of runtime. Second, the need to evaluate the unlabeled instances selected to train the algorithm. This evaluation process, that is, the fact of finding the real class of the selected unlabeled examples, is a fundamental point of the proposed classification model. This phase could affect the efficiency and accuracy of the model in general.

Table 3: Percentage of correctly classified instances (accuracy) and runtime of the studied algorithms on the real datasets.

| Bases/ Algorithms | SPA | WEA | ELE | CON | POK |
|---|---|---|---|---|---|
| Real Bases | | | | | |
| **FASE-AL** | **96,40** | **72,05** | 73,36 | **71,91** | **66,86** |
| | 8,03 | 0,66 | 1,40 | 5,16 | 9,30 |
| **Fixed-Uncertainty** | 95,20 | 71,40 | 48,40 | 60,43 | 42,70 |
| | **3,11** | 0,14 | 0,34 | 2,12 | **6,16** |
| **Var-Uncertainty** | 96,40 | 71,85 | 72,78 | 70,60 | 59,96 |
| | 3,14 | 0,12 | 0,28 | **2,08** | 8,16 |
| **Rand-Uncertainty** | 96,10 | 67,25 | 72,38 | 65,91 | 60,93 |
| | 3,12 | **0,11** | **0,27** | 2,11 | 8,03 |
| **Sel-Sampling** | 96,20 | 67,40 | **74,24** | 62,50 | 60,53 |
| | 3,22 | 0,12 | 0,30 | 2,17 | 9,17 |

## 6. CONCLUSIONS

In the present paper, the main characteristics of the algorithm ensemble called *FASE-AL* are detailed. *FASE-Al* is an adaptation of the known FASE algorithm to the active learning paradigm. This new algorithm is capable of learning online from a data stream with the presence of concept drift. Both synthetic data and real data were used in the design of the experiments. In the synthetic data, the presence of noise and the existence of abrupt and gradual concept drift were modeled.

For comparison with four other strategies, also designed to work on the active learning paradigm, two evaluation parameters used very frequently were taken into account: the percentage of correctly classified instances and the execution time. For the first of these parameters, *FASE-AL* achieved very promising results on both real and synthetic datasets. On the other hand, the results of the new algorithm were less promising when analyzing the execution time.

We propose as future work to conduct a study that analyzes the amounts and percentages of unlabeled instances necessary to keep the proposed algorithm stable. Analyze the behavior of the algorithm when the number of unlabeled instances is minimal, that is, find a minimum quantity or percentage.

## ACKNOWLEDGEMENTS

This work was partially funded by the Coordination of Improvement of Higher-Level Personnel - CAPES, and the Foundation of Support for Research and Innovation of Santa Catarina State - FAPESC.

## REFERENCES

[1]    Y. Fu, X. Zhu, B. Li, "A survey on instance selection for active learning". Knowl Inf Syst; Springer Verlag. 35:249283, DOI 10.1007/s10115-012-0507-8, 2013.

[2]     A. Ortiz, G. Ramos, J. del Campo, I. Frías, R. Bueno, "Fast Adapting Ensemble: A New Algorithm for Mining Data Streams with Concept Drift". The Scientic World Journal; Special Issue on Research and Development of Advanced Computing Technologies, 2015.

[3]     A. Tsymbal, "The problem of concept drift: definitions and related work". Tech. Rep. TCD-CS-2004-15, Department of Computer Science, Trinity College, Dublin, Ireland, 2004.

[4]     J. Gama, I. Zliobaite, A. Bifet, M. Pechenizkiy, and A. Bouchachia, "A Survey on Concept Drift Adaptation". ACM Comput. Surv., 46(4):44:144:37, March 2014.

[5]     Y. Padmanabha, P. Viswanath, B. Eswara. " Semi supervised learning: a brief review" International Journal of Engineering &Technology, 7 (1.8), pp 81-85, 2018.

[6]     Z. Zhou, M. Li, "Semi-supervised learning by disagreement", Knowl Inf Syst 24(3):415439, 2010.

[7]     W. Xu, S. Zhao, Z. Lu, "Active Learning over Evolving Data Streams using Paired Ensemble Framework". 8th International Conference on Advanced Computational Intelligence Chiang Mai, Thailand. IEEE 978-1-4673-7782-9/16, 2016.

[8]     I. Frías, A. Verdecia, A. Ortiz, A. Carvalho, "Fast adaptive stacking of ensembles. Proceedings of the 2016 ACM Symposium on Applied Computing". SAC: Pages 929-934, 2016.

[9]     J. Gama, P. Medas, G. Castillo, P. Rodrigues, "Learning with drift detection", in Proceedings of the 17th SBIA Brazilian Symposium on Arti_cial Intelligence, pp. 286295, Sao Luis, Brazil, Sep-Oct, 2004.

[10]    W. Street, Y. Kim, "A streaming ensemble algorithm (SEA) for large-scale classification". Proceedings of the 7th ACM SIGKDD International Conference on Knowledge Discovery and Data Mining (KDD '01), pp. 377382, New York, USA, Aug, 2001.

[11]    M. Baena, J. del Campo, R. Fidalgo, A. Bifet, R. Gavalda, R. Morales, "Early Drift Detection Method". In 4th Int. Workshop on Knowledge Discovery from Data Streams, 2006.

[12]    I. Frías, G. Ramos, J. del Campo, A. Ortiz, R. Bueno, "Online and non-parametric drift detection methods based on Hoefdings bounds". IEEE Transactions on Knowledge and Data Engineering, 14(3):810823, 2015.

[13]    J. Gama, P. Pereira and G. Castillo, "Evaluating Algorithms that Learn from Data Streams". In Proc. 2009 ACM Symposium on Applied Computing, pages 14961500, 2009.

[14]    A. Culotta and A. McCallum, "Reducing labeling effort for structured prediction tasks". Proceedings of the 20th national conference on artificial intelligence (AAAI), pp 746751 2005.

[15]    A. Holub, P. Perona, Entropy-based active learning for object recognition. IEEE computer society conference on computer vision and pattern recognition workshop anchorage (CVPR), pp 18, 2008.

[16]    I. liobaite, A. Bifet, B. Pfahringer and G. Holmes. Active learning with drifting streaming data. IEEE Transactions on Neural Net Works and Learning Systems, VOL. 25, NO. 1, jan 2014.

[17]    F. Zheng and I. Geoffrey. A Comparative Study of Semi-naive Bayes Methods in Classification Learning. Proceedings of the Fourth Australasian Data Mining Workshop, pages 141-156. Sydney: University of Technology, 2005.

[18]    A. Bifet, G. Holmes, R. Kirkby, and B. Pfahringer, MOA: Massive online analysis. Journal of Machine Learning Research, vol. 11, pp. 16011604, June 2010.

[19]    B. Krawczyk and A. Cano. Adaptive Ensemble Active Learning for Drifting Data Stream. Proceedings of the Twenty-Eighth International Joint Conference on Artificial Intelligence (IJCAI-19), 2019.

**Authors**

**Agustín Alejandro Ortiz Díaz** is graduated in Bachelor of Computer Science from the University of East, Cuba, (2003); Master's degree in Computer Science from the Central University of Las Villas, Cuba, (2011); and Ph.D. degree in Computer Science from the University of Granada, Spain (2014). He is currently a postdoctoral researcher at the Department of Computer Sciences, University of State of Santa Catarina, Brazil. He was a full professor at the University of Granma, Cuba (2003-2017). His research interests and results include machine learning, data mining, online learning, database and tools for big data analytics.

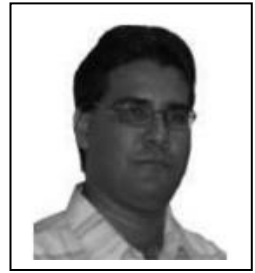

**Fabiano Baldo** is graduated in Bachelor of Computer Science from the Federal University of Santa Catarina (1999), Master's degree in Electrical Engineering from the Federal University of Santa Catarina (2003) and PhD in Electrical Engineering from the Federal University of Santa Catarina (2008). He is currently a professor at Santa Catarina State University and a researcher at Santa Catarina Federal University. Has experience in Computer Science, focusing on Information Systems, acting on the following subjects: Database, Systems Integration, Software Engineering.

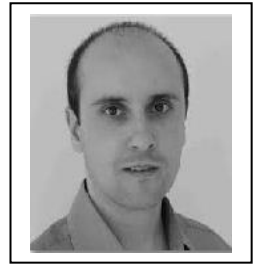

**Laura María Palomino Mariño** is a graduate in Computer Engineering from the University of Computer Science, UCI, Havana, Cuba (2007) and master's in computer science (CAPES Concept 7), Federal University of Pernambuco, UFPE, Brazil (2019). She is currently a student Ph.D. at the Federal University of Pernambuco, Brazil. He has experience in the areas of data mining and machine learning.

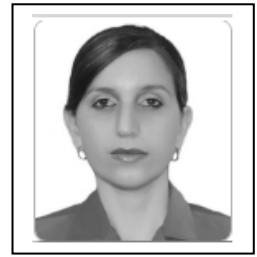

**Alberto Verdecia Cabrera** is a graduate in Computer Engineering from the University of Granma, Granma, Cuba (2010); and Ph.D. from the Central University of Las Villas, Santa Clara, Cuba. (2018). He is currently an assistant professor at the University of Granma, Cuba. He has experience in the areas of data mining, online classification, and machine learning.

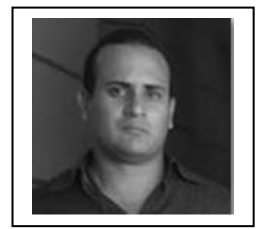